\pdfoutput=1

\documentclass[11pt]{article}

\usepackage[]{acl}

\usepackage{times}
\usepackage{latexsym}
\usepackage{amsmath}
\usepackage{graphicx}
\usepackage{booktabs}

\usepackage[T1]{fontenc}

\usepackage[utf8]{inputenc}

\usepackage{microtype}

\usepackage{inconsolata}

\title{HiCuLR: Hierarchical Curriculum Learning for Rhetorical Role Labeling of Legal Documents}

\author{ Santosh T.Y.S.S$^{1}$, \bf{Apolline Isaia$^{1,2}$, Shiyu Hong$^{1,3}$, Matthias Grabmair$^{1}$} \\ $^{1}$ Technical University of Munich, Germany \\ 
$^{2}$Télécom Paris \\
$^{3}$Future Technology School; South China University of Technology
}

\begin{document}
\maketitle
\begin{abstract}
Rhetorical Role Labeling (RRL) of legal documents is pivotal for various downstream tasks such as summarization, semantic case search and argument mining. Existing approaches often overlook the varying difficulty levels inherent in legal document discourse styles and rhetorical roles. In this work, we propose HiCuLR, a hierarchical curriculum learning framework for RRL. It nests two curricula: Rhetorical Role-level Curriculum (RC) on the outer layer and Document-level Curriculum (DC) on the inner layer. DC categorizes documents based on their difficulty, utilizing metrics like deviation from a standard discourse structure and exposes the model to them in an easy-to-difficult fashion. RC progressively strengthens the model to discern coarse-to-fine-grained distinctions between rhetorical roles. Our experiments on four RRL datasets demonstrate the efficacy of HiCuLR, highlighting the complementary nature of DC and RC. 
\end{abstract}

\section{Introduction}
Rhetorical Role Labeling (RRL) of legal documents involves assigning the functional role played by each sentence of the document in the legal discourse (e.g., preamble, fact, evidence, reasoning).
RRL for long legal case documents is a precursor task for several downstream tasks, such as case summarization \cite{hachey2006extractive,saravanan2008automatic,kalamkar2022corpus,farzindar2004letsum}, semantic case search \cite{nejadgholi2017semi,ali2021prior}, case law analysis \cite{savelka2018segmenting} and argument mining \cite{walker2019automatic,ali2023legal}.

Initial works on RRL of legal judgements have regarded the task as straightforward classification of sentences without modeling any contextual dependency between them \cite{walker2019automatic} which later got to be viewed as sequence labeling \cite{bhattacharya2021deeprhole}. Initial works \cite{saravanan2008automatic,farzindar2004letsum,hachey2006extractive} performed RRL using hand-crafted features as part of a summarization pipeline. Further works \cite{walker2019automatic,savelka2018segmenting} used Conditional Random Fields on these hand-crafted features. Recently, deep learning-based methods have been applied to this task on Japanese documents \cite{yamada2019neural}, Indian documents \cite{bhattacharya2021deeprhole,ghosh2019identification,malik2022semantic,kalamkar2022corpus}. These methods adopt a hierarchical approach to account for the sequential sentence classification nature of the task, drawing context from surrounding sentences. This architecture, initially based on word embeddings \cite{bhattacharya2021deeprhole,ghosh2019identification}, has transitioned to BERT-based contextualized embeddings \cite{malik2022semantic,kalamkar2022corpus}, becoming the standard approach for RRL tasks. Recently, \cite{santosh2023joint} reformulated the task as span-level sequential classification that segment the document into sets of contiguous sequence of sentences (spans) and assign them labels. Further \cite{tyss2024mind} proposed contrastive and prototypical learning to effectively leverage knowledge from semantically similar instances (neighbours). 

All these current approaches typically perform fine-tuning by presenting all training examples in a completely random order during training. However, the difficulty levels of discourse structures in legal documents can vary significantly, with some following a standard format identifiable by simple lexical clues, while others require deeper analysis due to unconventional writing styles. Similarly, identifying rhetorical roles such as preamble, which encompass broad definitions, is easier compared to finer-grained roles like the ratio of the decision, which is often confused with analysis or ruling by the present court. 
In this work, we leverage these observations to employ curriculum learning (CL) \cite{bengio2009curriculum}, where the model's training process mimics a meaningful learning order inspired by human cognition. By excluding the negative impact of difficult examples in the early training stages, CL can guide learning towards a better local minima in the parameter space \cite{hacohen2019power}. CL has demonstrated success in various NLP tasks such as machine translation \cite{kocmi2017curriculum,zhang2018empirical,zhou2020uncertainty,platanios2019competence}, NLU \cite{xu2020curriculum,maharana2022curriculum}, AMR Parsing \cite{wang2022hierarchical}, summarization \cite{kano2021quantifying,sotudeh2022curriculum,sun2023data}, data-to-text generation \cite{chang2021does}, information retrieval \cite{su2021dialogue,zhu2022easy,zeng2022curriculum}, pre-training \cite{ranaldi2023modeling,nagatsuka2021pre} and even in legal tasks such as statutory retrieval for mining negatives \cite{santosh2024cusines}.

In this work, we propose a hierarchical curriculum learning approach that nests two complementary curricula: (i) Document-level curriculum, which orders input documents based on their difficulty. We explore various difficulty metrics, such as deviations from standard discourse structure, to order these documents. We employ a baby-step scheduler \cite{spitkovsky2010baby}, gradually exposing the model to more difficult samples over time. (ii) Rhetorical role-level curriculum, which utilizes similarity-based scheduling \cite{dogan2020label}. Ordering rhetorical roles in a sequential easy-to-hard sequence, as in the document-level curriculum, is infeasible since we cannot disentangle sentences from a document, which is our input. Hence we use a similarity-based curriculum where the model is initially allowed to belong to similar rhetorical roles to a lesser extent, instead of solely belonging to its ground-truth role and it gets refined over training. That is, in the beginning small mistakes of similar classes are less corrected than bigger mistakes, resembling a teaching process in which broad concepts are explained before subtle differences are addressed later. Our experimental results on four RRL datasets demonstrate the effectiveness of our various difficulty metrics for both DC and RC, as well as our model-agnostic HiCuLR framework, which incorporates both curricula.

\section{Preliminaries}
\noindent \textbf{Task :} Given a judgment document $x = \{x_1,x_2,\ldots,x_m\}$ with m sentences as the input, where $x_i = \{x_{i1},x_{i2},\ldots,x_{in}\}$ represents the $i^\text{th}$ sentence containing $n$ tokens, 
the task of RRL is to predict sequence of $l = \{l_1,l_2,\ldots,l_m\}$ where  $l_i$ is the rhetorical role corresponding to sentence $x_i$ and $l_i \in$ L, set of predefined rhetorical roles. 

\noindent \textbf{Baseline :} We use Hierarchical Sequential Labeling Network from prior works \cite{kalamkar2022corpus,malik2022semantic} and demonstrate effectiveness of our model-agnostic HiCuLR framework in conjunction with it. Initially, each sentence $x_i$ is encoded independently using a BERT model \cite{devlin2018bert} to derive token-level representations. 
These token-representations are passed through a Bi-LSTM layer \cite{hochreiter1997long}, followed by an attention pooling layer \cite{yang2016hierarchical}, to yield sentence representations. 
These are passed through Bi-LSTM layer to obtain contextualized sentence representations 
to encode information from surrounding sentences 
which are passed through a CRF layer \cite{lafferty2001conditional} that predicts the best sequence of labels. 
The model is trained end-to-end with standard cross entropy loss.

\section{HiCuLR framework}
A curriculum learning framework consists of two main components - Difficulty scoring function which quantifies the data based on relative easiness and a pacing function which arranges the transitioning of data from easy to difficult examples for training  \cite{bengio2009curriculum}. In this section, we describe our document-level difficulty estimators along with the baby-step scheduler (\ref{DC}) and rhetorical role-level curriculum with similarity based scheduling (\ref{RC}). Further, we describe combining both of these complementary curricula in a nested hierarchical fashion 
(\ref{combining}).



\subsection{Document-level Curriculum (DC)}
\label{DC} 
We investigate four different strategies to obtain the relative difficulty across documents.

\noindent \emph{(a) Rhetorical Shifts :} We hypothesize that documents with more consecutive shifts in rhetorical roles are harder to comprehend. We calculate the difficulty score as the number of consecutive shifts of rhetorical roles in a document normalized by the number of sentences in that document.

\noindent \emph{(b) Deviation from Expert Discourse:} 
While there is no universally agreed-upon guideline for the discourse structure (sequence of rhetorical roles) in legal judgments, experts suggest an inherent structural pattern that is typically followed. Utilizing an expert-provided discourse structure\footnote{Expert-given discourse structure is obtained from Build respository at \url{https://github.com/Legal-NLP-EkStep/rhetorical-role-baseline}}, we posit that documents deviating more from this structure are more challenging. We quantify the deviation based on the number of inversions required in the merge sort algorithm to align the document's rhetorical role structure with the expert-provided one.

\noindent \emph{(c) Deviation from Data-based Best Discourse:} Instead of relying on expert-provided discourse structure, we derive the best structure by computing transition matrix for every pair of rhetorical roles using the training data for each dataset. We calculate the deviation score based on this sequence.

\noindent \emph{(d) Data-based Probabilistic Discourse:} We utilize the transition matrix probabilities to compute the difficulty score as the log-likelihood of the sequence of labels for each document, which is then normalized by length. Lower log-likelihood scores indicate greater difficulty.

Based on the difficulty score, we partition the dataset into buckets and adopt baby-step scheduling for training \cite{spitkovsky2010baby}. Initially, data from the easiest bucket is used and subsequent buckets are merged after a fixed number of steps, until the entire dataset is utilized.

\subsection{Rhetorical Role-level Curriculum (RC)}
\label{RC}
Given that the RRL task operates on entire documents as input, sequentially exposing the model only to sentences with easy rhetorical roles followed by difficult ones in a sequential fashion becomes impractical. To address this, we employ a similarity-based scheduler \cite{dogan2020label} for RC. This approach relies on a pairwise similarity matrix between rhetorical labels, which we obtain using two different strategies.

\noindent \emph{(a) Confusion Matrix:} We hypothesize that more mistakes a model makes between pairs of rhetorical roles, the more similar and confusing they are to the model. Therefore, we obtain the confusion matrix from the validation dataset by using the trained model in a non-curriculum (random) order as our pairwise similarity matrix.

\noindent \emph{(b) Embedding Similarity:} We pass the rhetorical role and their descriptions\footnote{Descriptionsare obtained from \url{https://github.com/Legal-NLP-EkStep/rhetorical-role-baseline}} to obtain the embeddings from the LegalBERT model \cite{chalkidis2020legal} to compute the semantic similarity between the rhetorical roles, resulting in a pairwise matrix.

The similarity-based curriculum \cite{dogan2020label} employs a probability distribution over rhetorical roles as the target label, unlike traditional one-hot encoding. This allows each sentence to be associated with similar rhetorical roles to a lesser extent, rather than being solely assigned to its ground-truth role. The class probabilities are initialized using a normalized similarity matrix to reflect the closeness among classes, and each row corresponds to a rhetorical label. This matrix is used in place of target probabilities when computing the cross-entropy loss.
As the training progresses, this similarity matrix is gradually shifted towards the standard one-hot-encoding (diagonal matrix). Each element in similarity matrix is updated as: 
\begin{equation}
    v_{ij}^{t+1} = \begin{cases}
      \frac{1}{1+ \varepsilon \sum_{k \neq j}  v_{ik}^t} & \text{$i = j$}\\
      \frac{\varepsilon v_{ij}^t}{1+ \varepsilon \sum_{k \neq j}  v_{ik}^t} & \text{$i \neq j$ }
    \end{cases}        
\end{equation}
$\varepsilon \in (0,1) $ controls the convergence rate of labels to one hot vectors. $v_{ij}^t$ represents probability of j-th element of i-th row at training step $t$. This approach aims to penalize major mistakes more heavily in the beginning, such as when the predicted class and target class are dissimilar, compared to minor mistakes when both the predicted and target class are similar. This resembles a teaching process where broader, easier rhetorical roles are learned before finer, closely associated rhetorical roles.

\subsection{Combining both Complementaries}
\label{combining}
Our HiCuLR framework hierarchically nests both of these curricula, with the rhetorical-level on the outside and the document-level curriculum on the inside. Within each step of the rhetorical role-level curriculum, the document-level curriculum is operated, starting from easy buckets and progressively expanding to the entire dataset, until the next step of the rhetorical curriculum takes place and the entire procedure is repeated iteratively.

\section{Experiments \& Results}
\begin{table*}[]
\scalebox{0.82}{
\begin{tabular}{|clcccccccc|}
\hline
\multicolumn{1}{|l|}{}   & \multicolumn{1}{l|}{}                                    & \multicolumn{2}{c|}{\textbf{Build}}                                    & \multicolumn{2}{c|}{\textbf{Paheli}}                                   & \multicolumn{2}{c|}{\textbf{M-CL}}                                     & \multicolumn{2}{c|}{\textbf{M-IT}}                \\ \hline
\multicolumn{1}{|l|}{}   & \multicolumn{1}{l|}{}                                    & \multicolumn{1}{c|}{\textbf{Mac-F1}} & \multicolumn{1}{c|}{\textbf{Mic-F1}} & \multicolumn{1}{c|}{\textbf{Mac-F1}} & \multicolumn{1}{c|}{\textbf{Mic-F1}} & \multicolumn{1}{c|}{\textbf{Mac-F1}} & \multicolumn{1}{c|}{\textbf{Mic-F1}} & \multicolumn{1}{c|}{\textbf{Mac-F1}} & \textbf{Mic-F1} \\ \hline
\multicolumn{1}{|c|}{0}  & \multicolumn{1}{l|}{Baseline}                            & \multicolumn{1}{c|}{60.02}    & \multicolumn{1}{c|}{78.43}    & \multicolumn{1}{c|}{61.53}    & \multicolumn{1}{c|}{66.48}    & \multicolumn{1}{c|}{58.42}    & \multicolumn{1}{c|}{66.21}    & \multicolumn{1}{c|}{63.54}    & 68.15    \\ \hline
\multicolumn{1}{|c|}{1}  & \multicolumn{1}{l|}{Rhetorical Shifts}                   & \multicolumn{1}{c|}{61.06}    & \multicolumn{1}{c|}{79.94}    & \multicolumn{1}{c|}{63.14}    & \multicolumn{1}{c|}{68.15}    & \multicolumn{1}{c|}{60.24}    & \multicolumn{1}{c|}{67.73}    & \multicolumn{1}{c|}{65.20}     & 69.76    \\ 
\multicolumn{1}{|c|}{2}  & \multicolumn{1}{l|}{Dev. from Expert Disc.}              & \multicolumn{1}{c|}{60.78}    & \multicolumn{1}{c|}{79.77}    & \multicolumn{1}{c|}{62.77}    & \multicolumn{1}{c|}{67.86}    & \multicolumn{1}{c|}{59.06}    & \multicolumn{1}{c|}{66.87}    & \multicolumn{1}{c|}{63.48}    & 68.54    \\ 
\multicolumn{1}{|c|}{3}  & \multicolumn{1}{l|}{Dev. from Data-based Best Disc.}     & \multicolumn{1}{c|}{60.64}    & \multicolumn{1}{c|}{79.64}    & \multicolumn{1}{c|}{63.10}     & \multicolumn{1}{c|}{68.35}    & \multicolumn{1}{c|}{61.15}    & \multicolumn{1}{c|}{68.88}    & \multicolumn{1}{c|}{64.22}    & 69.47    \\ 
\multicolumn{1}{|c|}{4}  & \multicolumn{1}{l|}{Data-based Prob. Disc.}              & \multicolumn{1}{c|}{\emph{61.14}}    & \multicolumn{1}{c|}{\emph{80.17}}    & \multicolumn{1}{c|}{\emph{63.52}}    & \multicolumn{1}{c|}{\emph{69.13}}    & \multicolumn{1}{c|}{\emph{61.85}}    & \multicolumn{1}{c|}{\emph{69.46}}    & \multicolumn{1}{c|}{\emph{65.29}}    & \emph{70.20}     \\ \hline
\multicolumn{1}{|c|}{5}  & \multicolumn{1}{l|}{Confusion Matrix}                    & \multicolumn{1}{c|}{60.82}    & \multicolumn{1}{c|}{80.66}    & \multicolumn{1}{c|}{\emph{64.12}}    & \multicolumn{1}{c|}{\emph{69.27}}    & \multicolumn{1}{c|}{61.74}    & \multicolumn{1}{c|}{70.13}    & \multicolumn{1}{c|}{\emph{66.29}}    & 71.24    \\ 
\multicolumn{1}{|c|}{6}  & \multicolumn{1}{l|}{Embedding Similarity}                & \multicolumn{1}{c|}{\emph{61.44}}    & \multicolumn{1}{c|}{\emph{81.28}}    & \multicolumn{1}{c|}{63.26}    & \multicolumn{1}{c|}{68.24}    & \multicolumn{1}{c|}{\emph{62.27}}    & \multicolumn{1}{c|}{\emph{71.42}}    & \multicolumn{1}{c|}{65.98}    & \emph{71.49}    \\ \hline
\multicolumn{1}{|c|}{7}  & \multicolumn{1}{l|}{Hierarchical: RC (5), DC(4)}        & \multicolumn{1}{c|}{\textbf{63.61}}    & \multicolumn{1}{c|}{\textbf{82.11}}    & \multicolumn{1}{c|}{\textbf{65.28}}    & \multicolumn{1}{c|}{70.12}    & \multicolumn{1}{c|}{\textbf{63.15}}    & \multicolumn{1}{c|}{\textbf{72.08}}    & \multicolumn{1}{c|}{\textbf{69.42}}    & 71.98    \\ 
\multicolumn{1}{|c|}{8}  & \multicolumn{1}{l|}{Hierarchical: RC (6), DC(4)}        & \multicolumn{1}{c|}{62.77}    & \multicolumn{1}{c|}{81.82}    & \multicolumn{1}{c|}{64.96}    & \multicolumn{1}{c|}{\textbf{70.48}}    & \multicolumn{1}{c|}{62.86}    & \multicolumn{1}{c|}{71.46}    & \multicolumn{1}{c|}{68.75}    & \textbf{72.27}    \\ \hline
\multicolumn{1}{|c|}{9}  & \multicolumn{1}{l|}{Sequential: DC(4), RC (5)}          & \multicolumn{1}{c|}{61.39}    & \multicolumn{1}{c|}{81.47}    & \multicolumn{1}{c|}{64.02}    & \multicolumn{1}{c|}{69.35}    & \multicolumn{1}{c|}{62.16}    & \multicolumn{1}{c|}{\emph{71.65}}    & \multicolumn{1}{c|}{\emph{68.12}}    & \emph{72.06}    \\ 
\multicolumn{1}{|c|}{10} & \multicolumn{1}{l|}{Sequential: RC(5), DC (4)}          & \multicolumn{1}{c|}{61.26}    & \multicolumn{1}{c|}{81.52}    & \multicolumn{1}{c|}{63.56}    & \multicolumn{1}{c|}{69.10}     & \multicolumn{1}{c|}{61.41}    & \multicolumn{1}{c|}{70.22}    & \multicolumn{1}{c|}{66.59}    & 71.92    \\ 
\multicolumn{1}{|c|}{11} & \multicolumn{1}{l|}{Reverse hier.: DC(4), RC (5)} & \multicolumn{1}{c|}{\emph{62.44}}    & \multicolumn{1}{c|}{\emph{81.84}}    & \multicolumn{1}{c|}{\emph{64.42}}    & \multicolumn{1}{c|}{\emph{69.72}}    & \multicolumn{1}{c|}{\emph{62.88}}    & \multicolumn{1}{c|}{71.47}    & \multicolumn{1}{c|}{67.21}    & 71.48    \\ \hline
\end{tabular}}
\caption{Performance comparison of various curriculum-based methods.  Dev., Disc., hier. indicate deviation, disourse and hierarchical respectively. Numbers in bracket for entries (7-11) denote the index number in first column, indicating the specific variant. Overall best and best in each sub-group are bolded and italicized respectively.}
\label{results_tab}
\end{table*}
We experiment on four RRL datasets: (i) Build \cite{kalamkar2022corpus},  (ii) Paheli \cite{bhattacharya2021deeprhole}
(iii) M-CL and (iv) M-IT \cite{malik2022semantic}, derived from Indian legal judgments. Build has 13 rhetorical roles annotated, while the others have 7 each. Detailed dataset descriptions and implementation details are in App. \ref{dataset} and \ref{impl}. Table \ref{results_tab} reports macro-F1 and micro-F1 scores.

\subsection{Results}
\textbf{Document-level Curriculum (DC):} Our analysis reveals that all four variants of difficulty scoring in DC led to improvements compared to the baseline across all datasets, highlighting the efficacy of using DC for RRL. Particularly, the data-based probability method (4) consistently outperformed other DC methods. While the deviation from expert discourse (2) performed slightly better or comparably to the data-based approach (3) on the Build dataset, its underperformance on other datasets suggests that the discourse structure derived from the Build annotation experts may not generalize well across different datasets. This suggests that a data-driven structure is a better proxy for expert-given discourse, facilitating easier adoption. Notably, the simple metric of rhetorical shifts (1), which does not incorporate specific label information, yielded better results than deviation-based methods (2, 3) in three out of four datasets, indicating its effectiveness as a strong signal for capturing document difficulty. This also signals the underutilization of the label information in the best sequence methods (2, 3), which is captured effectively using the data-based probabilistic discourse method (4). It suggests that using probabilistic method (4) facilitates the variability of possible discourse sequence styles as opposed to single best sequence in (2, 3). 

\noindent \textbf{Rhetorical Role-level Curriculum (RC):} Both the RC methods perform better than the baseline. However, there is no clear winner between them. We attribute this to the differences in the label space of the datasets - Build dataset has more labels compared to others, indicating their fine-grained nature, which is captured effectively by embedding similarity. Overall, RC performs slightly better than DC, indicating that the ordering of output labels is more important than that of input documents.

\noindent \textbf{HiCuLR:} We use the data-based probabilistic method (4) from DC and vary both the RC methods to create two HiCuLR variants (7,8). Overall, HiCuLR shows improvement compared to DC and RC alone, highlighting their complementary nature. Between them, confusion matrix takes the lead on challenging macro-F1. We also observe a trend reversal in HiCuLR compared to RC; for example, in the Build dataset, embedding similarity (6) in RC performs better, but within the HiCuLR, the confusion matrix (7) takes the lead. While embedding similarity reflects the label definition without considering the input, the confusion matrix captures the interplay of inputs and the associated rhetorical roles, making it a better proxy for overall dataset.

\noindent \textbf{Ablation on combining DC and RC:} We experiment with other strategies to combine DC and RC. Sequential strategy executes these curricula in a pipeline manner. Variant (9) executes DC first, followed by RC. DC-RC (9) outperforms RC-DC (10), suggesting that similarities between rhetorical roles towards the end of learning improve model performance. Additionally, combining both curricula in reverse hierarchical order, with RC on the outside and DC on the inside, performs better than sequential versions (9, 10), highlighting curricula interaction. Variant (11) underperforms compared to HiCuLR (7), indicating gradual exposure to rhetorical roles is preferable.

\section{Conclusion}
In this work, we proposed a novel hierarchical curriculum learning framework, HiCuLR, for RRL of legal documents. HiCuLR integrates complementary curricula, with Rhetorical Role-level Curriculum (RC) on the outer layer and Document-level Curriculum (DC) on the inner layer We investigate different difficulty scoring metrics in DC and similarity based strategies in RC, observing notable improvements in performance. Particularly, the data-based probabilistic method within DC and the confusion matrix approach within RC stood out as effective strategies. Our experiments on four RRL datasets, verify the effectiveness of HiCuRL.

\section*{Limitations}
One limitation of the current task setup is its constraint to assign single label per sentence, which may not fully account for the complexity of lengthy sentences that can encompass multiple rhetorical roles. To overcome this constraint, one alternative could be to rethink the task as a multi-label classification, enabling each sentence to be associated with more than one rhetorical role. Another avenue worth exploring is to move away from sentence-level segmentation towards a more detailed approach at the phrase or sub-sentence level. This would involve assigning rhetorical roles to individual phrases or sub-sentences \cite{tokala2023label} and also specifying the dependency relationships among these segments, similar to Discourse Dependency Parsing \cite{carlson2003building}.

Additionally, our evaluation is confined to datasets containing Indian legal documents. These datasets may share common vocabulary and writing style specific to the country's legal practices, potentially limiting the generalizability of our findings to legal documents from other jurisdictions. Since legal documents from different countries and regions may exhibit significant variations in language and structure, it's essential to broaden the assessment to include diverse legal contexts across different countries and regions. 

\section*{Ethics Statement}
The scope of this study is to introduce technical methodologies and corresponding empirical validations aimed at advancing rhetorical role labeling, a fundamental task at the forefront of legal document processing.  Our experiments have been conducted on four datasets sourced from various Indian courts, made available through earlier works. Whenever applicable, consent for data usage was obtained according to the terms and conditions provided by the dataset providers. While these datasets contain real names of involved parties and lack anonymization, we anticipate no adverse effects resulting from our experimentation. We assert that our research makes a constructive contribution to the overarching objectives of advancing legal NLP and fostering the creation of AI-driven tools to improve productivity of legal professionals. 

\bibliography{custom}

\appendix
\section{Dataset}
\label{dataset}
We experiment on four RRL datasets: 

\noindent \textbf{(i) Build} \cite{kalamkar2022corpus} includes judgments from Indian supreme court, high court and district courts, with publicly available train and validation splits. It comprises 184 and 30 documents respectively, totaling 31865 sentences, average of 115 per document, annotated with 13 rhetorical role labels. Due to the absence of a public test dataset, we use the training dataset for both training and validation, evaluating performance on the validation partition. 

\noindent \textbf{(ii) Paheli} \cite{bhattacharya2021deeprhole} features 50 judgments from the Supreme Court of India across five domains, with 7 rhetorical roles annotated. It contains 9380 sentences (average of 188 per document). We split into 80\% train, 10\% validation, and 10\% test set at the document level. 

\noindent \textbf{(iii) M-CL / (iv) M-IT} \cite{malik2022semantic} encompasses judgments from the Supreme Court of India, High Courts, and Tribunal courts, with two subsets: M-CL, comprising 50 documents related to Competition Law, and M-IT, with 50 documents related to Income Tax cases. Both subsets are annotated with 7 rhetorical role labels. M-CL has 13,328 sentences (average of 266 per document) 
and M-IT has 7856 sentences (average of 157 per document). We split M-CL / M-IT into 80\% train, 10\% validation, and 10\% test set at the document level. 

\section{Implementation Details}
\label{impl}
We follow the hyperparameters for baseline as described in \citealt{kalamkar2022corpus}. We use the BERT base model to obtain the token encodings. We employ a dropout of 0.5, maximum sequence length of 128, LSTM dimension of 768, attention context dimension of 200. We sweep over learning rates \{1e-5, 3e-5, 5e-5. 1e-4, 3e-4\} for 40 epochs with Adam optimizer \cite{kingma2014adam}. For HiCuLR, we vary interval step size for target similarity matrix updates in RC over $\{5,10,15,20,25\}$, decay factor $\varepsilon$ in RC over $\{0.8, 0.9, 0.95, 0.99, 0.999\}$,
number of difficulty buckets in DC over $\{3, 5, 7, 10, 12, 15\}$, max training epochs during each baby step in DC over $\{2, 4, 6, 8, 10\}$. We derive the best model based on validation set performance.

\end{document}